%% file: ms.tex
\documentclass[runningheads]{llncs}
\usepackage{graphicx}
\usepackage{cite}
\usepackage{amsmath,amssymb,amsfonts}
\usepackage{algorithm}
\usepackage[noend]{algpseudocode}

\usepackage{textcomp}
\usepackage{url} 
\usepackage{bbm}
\usepackage{calrsfs}
\usepackage{color}
\usepackage{multirow}
\usepackage{tikz}

\usepackage[utf8x]{inputenc}

\usepackage{caption}
\usepackage[framemethod=TikZ]{mdframed}
\usepackage{pgfplots}
\usepackage{colortbl}
\usepackage{xcolor}
\usepackage{comment}
\usepackage[export]{adjustbox}
\definecolor{snm}{rgb}{0,0,0}

\usepackage{pgfplots}
\pgfplotsset{compat=1.8}
\usepgfplotslibrary{statistics}

\begin{document}


\title{Multimodal Generalized Zero Shot Learning For Gleason Grading Using Self-Supervised Learning}

%

\author{Dwarikanath Mahapatra}


\authorrunning{D. Mahapatra et. al.}


\institute{Inception Institute of Artificial Intelligence, UAE \email{dwarikanath.mahapatra@inceptioniai.org} 
}

\maketitle

\begin{abstract}
Gleason grading from histopathology images is essential for accurate prostate cancer (PCa) diagnosis. Since such images are obtained after invasive tissue resection  quick diagnosis is challenging under the existing paradigm. We propose a method to predict Gleason grades from magnetic resonance (MR) images which are non-interventional and easily acquired. We solve the problem in a generalized zero-shot learning (GZSL) setting since we may not access training images of every disease grade. Synthetic MRI feature vectors of unseen grades (classes) are generated by exploiting Gleason grades' ordered nature through a conditional variational autoencoder (CVAE) incorporating self-supervised learning.  Corresponding histopathology features are generated using cycle GANs, and combined with MR features to predict Gleason grades of test images. Experimental results show our method outperforms competing feature generating approaches for GZSL, and comes close to performance of fully supervised methods.



\end{abstract}


\input{DGen_Intro}

\input{DGen_Method_2}

\input{DGen_Expt}

\input{DGen_Concl}

\bibliographystyle{splncs04}
\bibliography{ms}

\end{document}

%% file: DGen_Intro.tex
\section{Introduction}
\label{sec:intro}

Early and accurate diagnosis of prostate cancer (PCa) is an important clinical problem. 
High resolution histopathology images provide the gold standard but involve invasive tissue resection. Non-invasive techniques like magnetic resonance imaging (MRI) are useful for early abnormality detection \cite{Promise12,MonusacTMI,Mahapatra_Thesis,KuanarVC,MahapatraTMI2021,JuJbhi2020,Frontiers2020,Mahapatra_PR2020,ZGe_MTA2019,Behzad_PR2020} and easier to acquire, but their low resolution and noise makes it difficult to detect subtle differences between benign conditions and cancer. A combination of MRI and histopathology features can potentially leverage their respective advantages for improved accuracy in early PCa detection.
Accessing annotated multimodal data of same patient from same examination instance is a challenge. 
  Hence a machine learning model to generate one domain's features from the other is beneficial to combine them for PCa detection. 
Current supervised learning \cite{BhattaMiccai20,Mahapatra_CVIU2019,Mahapatra_CMIG2019,Mahapatra_LME_PR2017,Zilly_CMIG_2016,Mahapatra_SSLAL_CD_CMPB,Mahapatra_SSLAL_Pro_JMI,Mahapatra_LME_CVIU,LiTMI_2015,MahapatraJDI_Cardiac_FSL,Mahapatra_JSTSP2014,MahapatraTIP_RF2014}, and multiple instance learning \cite{GGL26,MahapatraTBME_Pro2014,MahapatraTMI_CD2013,MahapatraJDICD2013,MahapatraJDIMutCont2013,MahapatraJDIGCSP2013,MahapatraJDIJSGR2013,MahapatraTrack_Book,MahapatraJDISkull2012,MahapatraTIP2012,MahapatraTBME2011,MahapatraEURASIP2010,MahapatraTh2012,MahapatraRegBook} approaches for Gleason grading use all class labels in training. 
 Accessing labeled samples of all Gleason grades is challenging, and we also encounter known and unknown cases at test time, making the problem one of generalized zero-shot learning (GZSL). 
 We propose to predict Gleason grades for PCa  by generating histopathology features from MRI features and combining them for GZSL based disease classification.
GZSL classifies natural images from seen and unseen classes \cite{FelixEccv18,Mahapatra_CVAMD2021,PandeyiMIMIC2021,SrivastavaFAIR2021,Mahapatra_DART21b,Mahapatra_DART21a,LieMiccai21,TongDART20,Mahapatra_MICCAI20,Behzad_MICCAI20,Mahapatra_CVPR2020,Kuanar_ICIP19,Bozorgtabar_ICCV19,Xing_MICCAI19} and uses Generative Dual Adversarial Network (GDAN) \cite{HuangCVPR19,Mahapatra_ISBI19,MahapatraAL_MICCAI18,Mahapatra_MLMI18,Sedai_OMIA18,Sedai_MLMI18,MahapatraGAN_ISBI18,Sedai_MICCAI17,Mahapatra_MICCAI17,Roy_ISBI17,Roy_DICTA16,Tennakoon_OMIA16,Sedai_OMIA16}, overcomplete distributions \cite{KeshariCvpr20,Mahapatra_OMIA16,Mahapatra_MLMI16,Sedai_EMBC16,Mahapatra_EMBC16,Mahapatra_MLMI15_Optic,Mahapatra_MLMI15_Prostate,Mahapatra_OMIA15,MahapatraISBI15_Optic,MahapatraISBI15_JSGR,MahapatraISBI15_CD,KuangAMM14,Mahapatra_ABD2014,Schuffler_ABD2014,Schuffler_ABD2014_2,MahapatraISBI_CD2014} and domain aware visual bias elimination \cite{MinCVPR20,MahapatraMICCAI_CD2013,Schuffler_ABD2013,MahapatraProISBI13,MahapatraRVISBI13,MahapatraWssISBI13,MahapatraCDFssISBI13,MahapatraCDSPIE13,MahapatraABD12,MahapatraMLMI12,MahapatraSTACOM12,VosEMBC,MahapatraGRSPIE12,MahapatraMiccaiIAHBD11,MahapatraMiccai11,MahapatraMiccai10}. But it has not been well explored for medical images. One major reason being the availability of class attribute vectors for natural images that describe characteristics of seen and unseen classes, but are challenging to obtain for medical images. 
Self-supervised learning (SSL) also addresses labeled data shortage and has found wide use in medical image analysis by using innovative pre-text tasks for active learning \cite{MahapatraTMI2021,MahapatraICIP10,MahapatraICDIP10a,MahapatraICDIP10b,MahapatraMiccai08,MahapatraISBI08,MahapatraICME08,MahapatraICBME08_Retrieve,MahapatraICBME08_Sal,MahapatraSPIE08,MahapatraICIT06,Covi19_Ar,DARTSyn_Ar,Kuanar_AR2,TMI2021_Ar,Kuanar_AR1}, anomaly detection \cite{Behzad_MICCAI20,Lie_AR2,Lie_AR,Salad_AR,Stain_AR,DART2020_Ar,CVPR2020_Ar,sZoom_Ar,CVIU_Ar,AMD_OCT,GANReg2_Ar,GANReg1_Ar,PGAN_Ar,Haze_Ar,Xr_Ar,RegGan_Ar}, and data augmentation \cite{Mahapatra_CVPR2020,ISR_Ar,LME_Ar,Misc,Health_p,Pat2,Pat3,Pat4,Pat5,Pat6,Pat7,Pat8,Pat9,Pat10}.
 SSL has been applied to histopathology images using domain specific pretext tasks \cite{SelfPath,Pat11,Pat12,Pat13,Pat14,Pat15,Pat16,Pat17,Pat18}, semi-supervised histology classification \cite{LuMahmood}, stain normalization \cite{Mahapatra_MICCAI20}, registration \cite{TongDART20} and cancer subtyping  using visual dictionaries.
Wu et al. \cite{WuCvpr20} combine SSL and GZSL for natural images but rely on class attribute vectors and unlabeled target data for training. 

Our current work makes the following contributions:
1) We propose a  \textbf{multimodal framework for seen and unseen Gleason grade prediction from MR images by combining GZSL and SSL.} 2) Unlike previous methods used for natural images we do not require class attribute vectors nor unlabeled target data during training.  
3) We propose a \textbf{self-supervised learning} (SSL) approach for feature synthesis of seen and unseen classes that exploits the ordered relationship between different Gleason grades to generate new features. 
 Although the method by \cite{BhattaMiccai20} uses features from MRI and histopathology images:1) it is a fully supervised method that has no Unseen classes during test time; 2) it also uses MRI and histopathology features from available data while we generate synthetic features to overcome unavailability of one data modality.

%% file: DGen_Method_2.tex
\section{Method}
\label{sec:method}

\begin{figure}[t]
 \centering
\begin{tabular}{c}
\includegraphics[height=5.5cm, width=10.6cm]{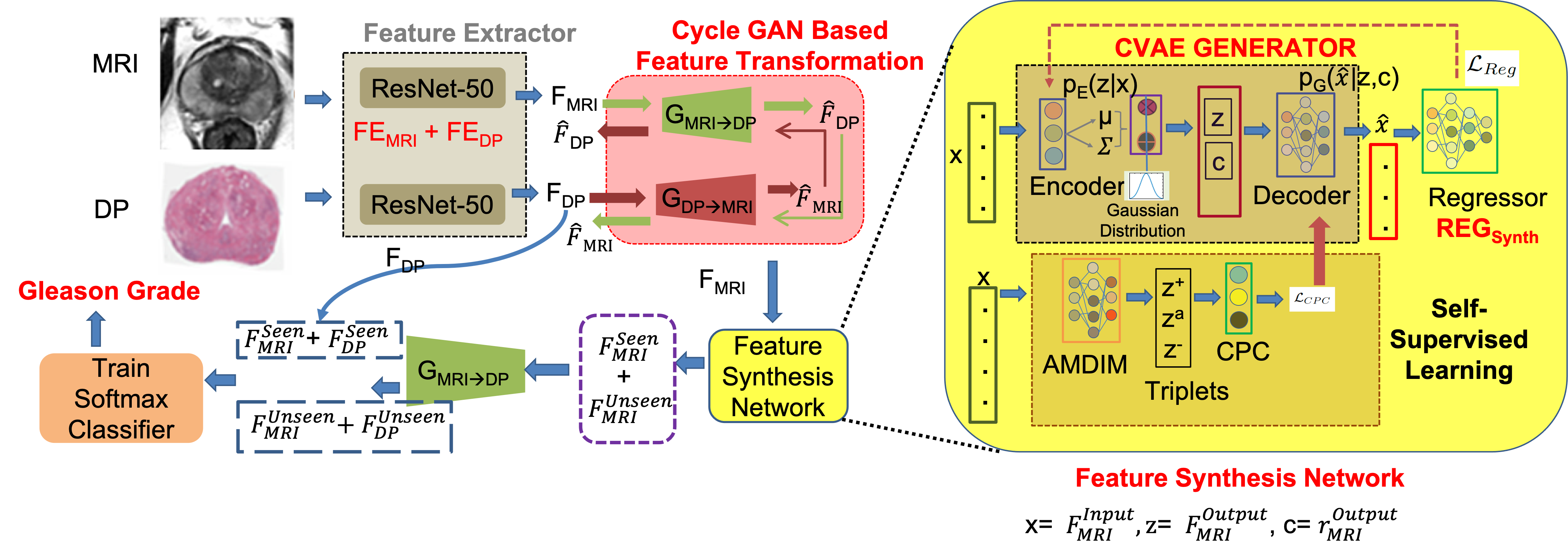} \\
\end{tabular}
 \caption{\textbf{Training Workflow:} Feature extraction from MR and digital pathology images generates respective feature vectors $F_{MRI}$ and $F_{DP}$. A $F_{MRI}$ is input to a feature synthesis network to generate new MR features for `Seen' and `Unseen' classes, which are passed through the feature transformation network to obtain corresponding $F_{DP}$. Combined MR and histopathology features are used to train a softmax classifier for identifying `Seen' and `Unseen' test classes. A self-supervised module contributes to the sample generation step through $\mathcal{L}_{CPC}$.
 }
\label{fig:workflow}
\end{figure}

\subsection{Feature Extraction And Transformation:}
\label{met:feat}

Figure~\ref{fig:workflow} depicts our overall proposed workflow, which has 3 networks for: feature extraction, feature transformation and self supervised feature synthesis. 
Let the training set consist of \emph{histopathology and MR images} from PCa cases with known Gleason grades. We train a network that can correctly predict seen and unseen Gleason grades using \emph{MR images only}. %
Let the dataset of `Seen' and `Unseen' classes be $S,U$ and their corresponding labels be $\mathcal{Y}_{s},\mathcal{Y}_{u}$.  $\mathcal{Y}=\mathcal{Y}_{s} \cup \mathcal{Y}_{u}$, and $\mathcal{Y}_{s} \cap \mathcal{Y}_{u}=\emptyset$.
Previous works \cite{FelixEccv18} show that generating feature vectors, instead of images, performs better for GZSL classification since output images of generative models can be blurry, especially with multiple objects (e.g., multiple cells in histopathology images).
Additionally, generating high-resolution histopathology images from low resolution and noisy MRI is challenging, while finding a transformation between their feature vectors is much more feasible.

We train two separate ResNet-50 networks \cite{ResNet50} as feature extractors for MR and DP images.  
 Histopathology image feature extractor ($FE_{DP}$) is pre-trained  on the PANDA dataset \cite{Panda} that has a large number of whole slide images for PCa classification.  
 $FE_{MRI}$ is pre-trained with the PROMISE12 challenge dataset \cite{Promise12} in a self-supervised manner. Since  PROMISE12 is for prostate segmentation from MRI and does not have classification labels, we use a pre-text task of predicting if an image slice has the prostate or not. Gleason grades of MRI are the same as corresponding histopathology images. The images are processed through the convolution blocks and the $1000$ dimensional output of the last FC layer is the feature vector for MRI ($F_{MRI}$) and pathology images ($F_{DP}$).

\textbf{Feature Transformation} is achieved using 
%
 CycleGANs \cite{CyclicGANS,MahapatraGAN_ISBI18,Mahapatra_PR2020,Mahapatra_MLMI18,Mahapatra_ISBI19} to learn mapping functions $G : X \rightarrow Y$ and $F : Y \rightarrow X$,  between feature vectors $X=F_{MRI}$ and $Y=F_{DP}$. Adversarial discriminator $D_X$ differentiates  between real features $F_{DP}$ and generated features $\widehat{F}_{DP}$, and $D_Y$ distinguishes between $F_{MRI}$ and $\widehat{F}_{MRI}$. 
The adversarial loss (Eqn~\ref{eqn:cyGan1}) and cycle consistency loss (Eqn.~\ref{eqn:cyGan2}) are, 
\begin{equation}
\begin{split}
L_{adv}(G,D_Y) = \mathbb{E}_{y} \left[\log D_Y(y)\right] + \mathbb{E}_{x} \left[\log \left(1-D_Y(G(x))\right)\right], \\
L_{adv}(F,D_X) = \mathbb{E}_{x} \left[\log D_X(x)\right] + \mathbb{E}_{y} \left[\log \left(1-D_X(F(y))\right)\right].
\end{split}
\label{eqn:cyGan1}
\end{equation}
\begin{equation}
L_{cyc}(G,F)= E_{x} \left\|F(G(x))-x\right\|_1 + E_{y} \left\|G(F(y))-y\right\|_1.
\label{eqn:cyGan2}
\end{equation}

\paragraph{\textbf{Network Training}}:
 is done using 
 $L_{adv}(G,D_Y)+L_{adv}(F,D_X)+L_{cyc}(G,F)$. Generator $G$ is a multi-layer perceptron (MLP) with a hidden layer of $4096$ nodes having LeakyReLU \cite{Felix34}. The output layer with $2048$ nodes has ReLU activation \cite{Felix35}. $G$'s weights are initialized with a truncated normal of $\mu=0$, $\sigma=0.01$, and biases initialized to $0$. Discriminator $D$ is an MLP with a hidden layer of  $2048$ nodes activated by LeakyReLU, and the output layer has no activation. $D$'s initialization is the same as $G$, and we use Adam optimizer \cite{Adam}.

\subsection{CVAE Based Feature Generator Using Self Supervision}

The conditional variational autoencoder (CVAE) generator synthesizes feature vectors $F_{MRI}^{output}$ of desired class $c_{MRI}^{output}$, given input features $F_{MRI}^{input}$ with known class $c_{MRI}^{input}$. Let $x=F_{MRI}^{input}, z=F_{MRI}^{output}$ and $c=c_{MRI}^{output}$.
The CVAE loss is, 
\begin{equation}
    \min_{\theta_G,\theta_E} \mathcal{L}_{CVAE} + \lambda_c \cdot \mathcal{L}_c + \lambda_{reg} \cdot \mathcal{L}_{reg} + \lambda_E \cdot \mathcal{L}_E + \lambda_{CPC} \cdot \mathcal{L}_{CPC}
    \label{eq:lfinal}
\end{equation}

Denoting CVAE encoder as $p_E(z|x)$ with parameters $\theta_E$, and the regressor output distribution as $p_R(c|x)$, $\mathcal{L}_{CVAE}$ is: 
%
\begin{equation}
    \mathcal{L}_{CVAE}(\theta_E,\theta_G)= -\mathbb{E}_{p_E(z|x),p(c|x)}\left[\log p_G(x|z,c)\right] + KL(p_E(z|x)||p(z))
\label{eq:lcvae}
\end{equation}
$\mathbb{E}_{p_E(z|x),p(c|x)}(.)$ is the generator’s reconstruction error, and KL divergence, $KL(.)$, encourages CVAE posterior (the encoder) to be close to the prior. 
 Encoder $p_E(z|x)$, conditional
decoder/generator $p_G(x|z, c)$, and regressor $p_R(c|x)$ are modeled
as Gaussian distributions. The  latent code $(z, c)$ is represented by disentangled representations $p_E(z|x)$ and $p_R(c|x)$ to avoid posterior collapse \cite{Verma11} .

\subsubsection{Regression/Discriminator Module - $REG_{Synth}$:}
\label{syn:reg}

%
 %
is a feedforward neural network mapping input feature vector $x \in \mathcal{R}^{D}$ to its corresponding class-value $c \in \mathcal{R}^{1}$ . $REG_{Synth}$ is a probabilistic model $p_R(c|x)$ with parameters $\theta_R$ and is trained using supervised ($\mathcal{L}_{Sup}$) and unsupervised ($\mathcal{L}_{Unsup}$) losses:
\begin{equation}
    \min_{\theta_R} \mathcal{L}_R = \mathcal{L}_{Sup} + \lambda_R \cdot \mathcal{L}_{Unsup}
    \label{eq:lreg2}
\end{equation}
$\lambda_R=0.2$ and 
 $\mathcal{L}_{Sup}(\theta_R)=-\mathbb{E}_{ \{x_n,c_{n} \} }\left[ p_R(c_{n}|x_n) \right]$ is defined on labeled examples $\{x_n,c_{n}\}^{N}_{n=1}$ from the seen class.
$\mathcal{L}_{Unsup}(\theta_R)=-\mathbb{E}_{p_{\theta_G} (\widehat{x}|z,c)p(z)p(c)} \left[ p_R(c|\widehat{x}) \right]$ is defined on synthesized examples $\widehat{x}$ from the generator.
 $\mathcal{L}_{Unsup}$ is obtained by sampling $z$ from $p(z)$, and class-value $c$ sampled from $p(c)$ to generate an exemplar for $p_{\theta_G}(\widehat{x}|z,c))$, and we calculate the expectation w.r.t. these distributions.

\subsubsection{Discriminator-Driven Learning:}
The error back-propagated from $REG_{Synth}$ improves quality of synthetic samples $\widehat{x}$ making them similar to the desired output class-value $c$. This is achieved using multiple loss functions. The first  generates samples whose regressed class value is close to the desired value, %
\begin{equation}
 \mathcal{L}_c(\theta_G)=-\mathbb{E}_{p_G(\widehat{x}|z,c)p(z)p(c)}\left[\log p_R(c|\widehat{x}) \right]   
\end{equation}
The second term draws samples from prior $p(z)$ and combines with class-value from $p(c)$, to ensure the synthesized features are similar to the training data,
\begin{equation}
 \mathcal{L}_{Reg}(\theta_G)=-\mathbb{E}_{p(z)p(c)}\left[\log p_G(\widehat{x}|z,c) \right]   
 \label{eq:lreg}
\end{equation}

The third loss term ensures independence (disentanglement) \cite{Verma11} of $z$ from
the class-value $c$. The encoder 
ensures that the sampling distribution and the one obtained
from the generated sample follow the same distribution.
%
\begin{equation}
 \mathcal{L}_{E}(\theta_G)=-\mathbb{E}_{\widehat{x} \sim p_G (\widehat{x}|z,c)} KL\left[\log p_E(z|\widehat{x})||q(z) \right]   
 \label{eq:le}
\end{equation}
The distribution $q(z)$ could be the prior $p(z)$ or the posterior from a labeled sample $p(z|x_n)$. 

\subsubsection{Self Supervised Loss:}
%
%

%

Gleason grades have a certain ordering, i.e., grade 3 indicates higher severity than grade $1$, and grade $5$ is higher than grade $3$. Contrastive Predictive Coding (CPC) \cite{Oord} learns self-supervised representations by 
predicting future observations from past ones and requires that observations be ordered in some dimension. %
Inspired by CPC we train our network to predict features of desired Gleason grade from input features of a different grade. From the training data we construct pairs of $\{F_{MRI}^{input},c_{MRI}^{input},F_{MRI}^{output},c_{MRI}^{output}\}$, the input and output features and desired class label value $c_{MRI}^{output}$ of synthesized vector. 

Since the semantic gap between $F_{MRI}^{input}$ and $F_{MRI}^{output}$ may be too large, we use random transformations to first generate intermediate representation of positive $z^{+}$, anchor $z^{a}$ and negative $z^{-}$ features using the mutual information-based AMDIM approach of \cite{AMDIM}. AMDIM ensures that the mutual information between similar samples $z^{+},z^{a}$ is high while for dissimilar samples $z^{-},z^{a}$ it is low, where $z=F_{MRI}^{output}$.  %
The CPC objective (Eqn.~\ref{eq:lcpc}) evaluates the quality of the predictions  using a contrastive loss where the goal is to correctly recognize the synthetic vector $z$ among a set of randomly sampled feature vectors $z_l=\{z^{+},z^{a},z^{-}\}$.
\begin{equation}
        \mathcal{L}_{CPC}  
        = -\sum \log \frac{\exp^{(F_{MRI}^{input})^{T} F_{MRI}^{output}}} {\exp^{(F_{MRI}^{input})^{T} F_{MRI}^{output}} + \sum \exp^{(F_{MRI}^{input})^{T} F_{MRI}^{output}} }
    \label{eq:lcpc}
\end{equation}
This loss is the InfoNCE inspired from Noise-Contrastive Estimation \cite{Gutmann2010,Mnih2013} and maximizes  the mutual information between $c$ and $z$ \cite{Oord}.
%
%
%
%
%
%
%
By specifying the desired class label value of synthesized vector and using it to guide feature generation in the CVAE framework we avoid the pitfalls of unconstrained and unrealistic feature generation common in generative model based GZSL  (e.g.,\cite{Manc48,Felix8}).

\textbf{Training And Implementation:}
Our framework was implemented in PyTorch. The CVAE Encoder has two hidden layers of $2000$ and $1000$ units respectively while the CVAE Generator is implemented with one hidden layer of $1000$ hidden
units. The Regressor has only one hidden layer of $800$ units. We choose Adam \cite{Adam} as our optimizer, and the momentum is set to ($0.9, 0.999$).The learning rate for
CVAE and Regressor is $0.0001$.
 First the CVAE loss (Eqn.~\ref{eq:lcvae}) is pre-trained followed by joint training of regressor and encoder-generator pair in optimizing Eqn.~\ref{eq:lreg} and Eqn.~\ref{eq:lfinal} till convergence.  Hyperparameter values were chosen using a train-validation split with 
 $\lambda_R = 0.1, \lambda_c = 0.1, \lambda_{reg} = 0.1$, $\lambda_E = 0.1$, and $\lambda_{CPC} = 0.2$. Training the feature extractor for $50$ epochs takes 12 hours and the feature synthesis network for $50$ epochs takes 17 hours, all on a single NVIDIA V100 GPU (32 GB RAM).

\textbf{ Evaluation Protocol:}
%
The seen class $S$ has samples from $2$ or more Gleason grades, and the unseen class $U$ contains samples from remaining classes. $80-20$ split of $S$ is done into $S_{Train}/S_{Test}$.
 $F_{MRI}$ is synthesized from $S + U$ using the CVAE and combined with the corresponding synthetic  $F_{DP}$ to obtain a single feature vector for training a softmax classifier minimizing the negative log-likelihood loss.
Following standard practice for GZSL, average class accuracies are calculated for two settings:  1) $Setting ~\textbf{A}$: training is performed on synthesized samples of $S+U$ classes and test on $S_{Test}$. 2) $Setting ~\textbf{B}$: training is performed on synthesized samples of $S+U$ classes and test on $U$. Following GZSL protocol we  report the harmonic mean:
$H=\frac{2\times Acc_U \times Acc_S}{Acc_U + Acc_S}$;
 $Acc_S$ and $Acc_U$  are classification accuracy of images from seen (setting $A$) and unseen (setting $B$) classes respectively:

%% file: DGen_Expt.tex

\section{Experimental Results}
\label{sec:expt}

\subsubsection{Dataset Details:}

 We use images from $321$ patients ($48-70$ years; mean: $58.5$ years) undergone prostate surgery with pre-operative T2-w and Apparent Diffusion Coefficient MRI, and post-operative digitized histopathology images. To ensure prostate tissue is sectioned in same plane as T2w MRI custom 3D printed molds were used. Majority votes among three pathologists provided Gleason grades was the consensus annotation. Pre-operative MRI and histopathology images were registered \cite{Bhatta11} to enable accurate mapping of cancer labels.
 We have the following classes: Class $1: Grade-3, 67$ patients, Class $2:Grades-3+4/4+3, 60$ patients, Class $3:Grade-4, 74$ patients, Class $4:Grades-4+5/5+4, 57$ patients, Class $5:Grade-5, 63$ patients. In the supplementary we show results for the Kaggle DR challenge \cite{Kaggle}  and PANDA challenge \cite{Panda}.

\textbf{Pre-processing:}
Histopathology images were smoothed with a Gaussian filter ($\sigma= 0.25$). They were downsampled to $224 \times 224$ with X-Y resolution $0.25 \times 0.25$ mm$^{2}$. The T2w images, prostate masks, and Gleason labels are projected on the corresponding downsampled histopathology images and resampled to the same X-Y resolution. This ensures corresponding pixels in each modality represents the same physical area.
 MR images were normalized using histogram alignment \cite{Bhatta12}.%
 The training, validation, and test sets had $193/64/64$ patients.


We compare results of our method MM$_{GZSL}$ (Multimodal GZSL) with different feature generation based GZSL methods:
1) CVAE based feature synthesis method of \cite{HuangCVPR19}; 2) GZSL using over complete distributions \cite{KeshariCvpr20}; 3) self-supervised learning GZSL method of \cite{WuCvpr20}; 4) cycle-GAN GZSL method of \cite{FelixEccv18}; 5) $FSL$- the fully supervised learning method of \cite{BhattaMiccai20} using the same data split, actual labels of `Unseen' class and almost equal representation of all classes.  

 \begin{table}[ht]
\begin{center}
\begin{tabular}{|c|c|c|c|c|c|c|c|c|c|c|c|c|}
\hline  
{} & {$Acc_S$} & {$Acc_U$} & {H} & {p} & {$Sen_S$} & {$Spe_S$} & {$Sen_U$} & {$Spe_U$} \\ \hline
\multicolumn{9}{|c|}{\textbf{Comparison Methods}} \\ \hline
{MM$_{GZSL}$} & {83.6(2.4)} & {81.7(3.0)} & {82.6(2.8)} & {-} & {84.1(3.1)} & {82.9(2.6)} & {81.2(3.0)} & {80.1(3.3)}\\ \hline 
{\cite{HuangCVPR19}} & {80.3(3.5)} & {73.4(3.6)} & {76.7(2.8)} & {0.002} & {81.2(3.6)} & {79.9(3.5)}  & {74.1(3.2)}  & {72.8(3.4)} \\ \hline
\cite{KeshariCvpr20}  & {80.6(3.4)}  & {72.8(3.0)}  & {76.5(3.2)}  & {0.001} & {81.1(2.9)} & {80.0(3.2)} & {73.5(3.1)} & {72.1(3.4)}  \\ \hline
\cite{WuCvpr20} & {81.1(2.9)} & {73.2(3.2)} & {76.9(3.1)} & {0.001} & {81.8(3.5)}  & {80.7(3.1)} & {74.0(3.5)} & {72.9(3.7)} \\ \hline
\cite{FelixEccv18} & {81.2(3.7)} & {72.8(3.8)} & {76.7(3.8)} & {0.004} & {81.8(3.1)} & {80.7(3.4)} & {73.1(4.0)} & {71.9(4.2)} \\ \hline 
FSL & {83.9(2.2)} & {83.3(2.5)} & {83.5(2.3)} & {0.01}  & {84.9(2.4)} & {83.5(2.6)} & {83.7(2.8)} & {82.5(2.6)} \\ \hline
\multicolumn{9}{|c|}{\textbf{Ablation Studies}} \\ \hline
{MM$_{wCPC}$} & {79.1(3.1)} & {72.3(3.8)} & {75.5(3.4)} & {0.001} & {79.8(3.4)} & {78.3(3.5)} & {73(3.6)} & {82.5(3.1)}\\ \hline
{MM$_{wReg}$}  & {80.1(3.7)} & {72.2(3.5)}  & {75.9(3.5)} & {0.0001} & {80.5(3.4)}  & {78.8(3.5)} & {72.8(3.8)} & {71.3(4.0)} \\ \hline
{MM$_{wC}$} & {79.2(3.9)} & {72.9(4.0)} &  {75.9(3.9)} & {0.009} & {80.1(3.8)} & {78.6(4.3)} & {73.3(4.1)} & {72.1(4.2)} \\ \hline
{MM$_{wE}$} & {80.2(3.7)} & {73.0(3.9)} & {76.4(3.7)} & {0.005} & {80.8(3.3)} & {79.5(3.8)} & {74.1(3.8)} & {72.4(3.4)} \\ \hline
{MM$_{MR}$} & {75.6(4.2)} & {71.1(4.5)} & {73.3(4.3)} & {0.007}  & {76.3(4.5)} & {75.0(4.3)} & {71.9(4.1)}  & {70.4(4.3)}\\ \hline
{MM$_{DP}$} & {81.2(3.0)} & {79.1(3.6)} &  {80.1(3.4)} & {0.008} & {82.0(3.4)} & {80.6(3.5)} & {79.9(3.9)} & {78.4(4.2)}  \\ \hline
\end{tabular}
\caption{\textbf{GZSL and Ablation Results:} Average classification accuracy ($\%$) and harmonic mean accuracy of generalized zero-shot learning when test samples are from  Seen (Setting $A$) or unseen (Setting $B$) classes. Mean and variance are reported when the number of classes in the `Seen' set is $3$. $p$ values are with respect to \textbf{Harmonic Mean of MM$_{GZSL}$}.}
\label{tab:GZSL_3class}
\end{center}
\end{table}

\subsection{Generalized Zero Shot Learning Results}

Table \ref{tab:GZSL_3class} summarizes the results of our algorithm and other methods  when the Seen set has samples from $3$ classes. The numbers are an average of $5$ runs. Samples from  $3$ labeled classes presents the optimum scenario balancing  high classification accuracy, and  generation of representative synthetic samples. 
%
Setting $A$ does better than setting $B$. Since GZSL is a very challenging problem, it is expected that classification performance on Unseen classes will not match those of Seen classes. 
MM$_{GZSL}$'s performance is closest to FSL. Since FSL has been trained with all classes in training and test sets it gives the best results. %

We use the McNemar test  and determine that the difference in $Acc_S$ of MM$_{GZSL}$ and FSL is not significant ($p=0.062$) although the corresponding values for $Acc_U$ are significant ($p=0.031$) which is not surprising since real Unseen examples have not been encountered in GZSL. 
Since MM$_{GZSL}$'s performance is closest to FSL for Unseen classes, it  demonstrates MM$_{GZSL}$'s effectiveness in generating  realistic features of unseen classes.  
We refer the reader to the supplementary material for additional results (e.g. using different number of classes in the Seen dataset). 
%
The individual Per Class mean accuracy and variance are: \textbf{Setting~A:-}Class1=84.1(2.4), Class2=83.8(2.8), Class3=84.1(2.3), Class4=82.9(2.8), Class5=82.9(3.1). \textbf{Setting~B-} Class1=83.1(2.7), Class2=82.9 (2.8), Class3=81.1 (2.9), Class4=79.0(3.2), Class5=78.2(3.9). This shows that our feature generation and classification is not biased to any specific class. %

\subsubsection{Ablation Studies:}

%
Table~\ref{tab:GZSL_3class} shows ablation results where each row denotes a specific setting without the particular term in the final objective function in Eqn~\ref{eq:lfinal} - e.g.,  MM$_{wCPC}$ denotes our proposed method MM$_{GZSL}$ without the self-supervised loss $\mathcal{L}_{CPC}$. 
MM$_{wCPC}$ shows the worst performance indicating $\mathcal{L}_{CPC}$'s correspondingly higher contribution than other terms. The $p-$values indicate each term's contribution is significant  for the overall performance of MM$_{GZSL}$ and excluding any one leads to significant performance degradation. 

We also show in Table~\ref{tab:GZSL_3class} the result of using: 1) only the synthetic MR features (MM$_{MR}$); 2) only digital histopathology features (MM$_{DP}$) obtained by transforming $F_{MRI}$ to get $F_{DP}$. MM$_{DP}$ gives performance metrics closer to MM$_{GZSL}$, which indicates that the digital histopathology images provide highly discriminative information compared to MR images. However, MR images also have a notable contribution, as indicated by the p-values.
In another set of experiments we redesign the GZSL experiments to generate synthetic histopathology features $F_{DP}$ instead of $F_{MRI}$ and then transforming them to get MR features. We obtain very similar performance ($Acc_S=83.7, Acc_U=80.8, H=82.2$) to the original MM$_{GZSL}$ setting. %
 This demonstrates that cycle GAN based feature transformation network does a good job of learning accurate histopathology features from the corresponding MR features.

\begin{figure}[t]
 \centering
\begin{tabular}{cccc}
\includegraphics[height=2.3cm, width=3cm]{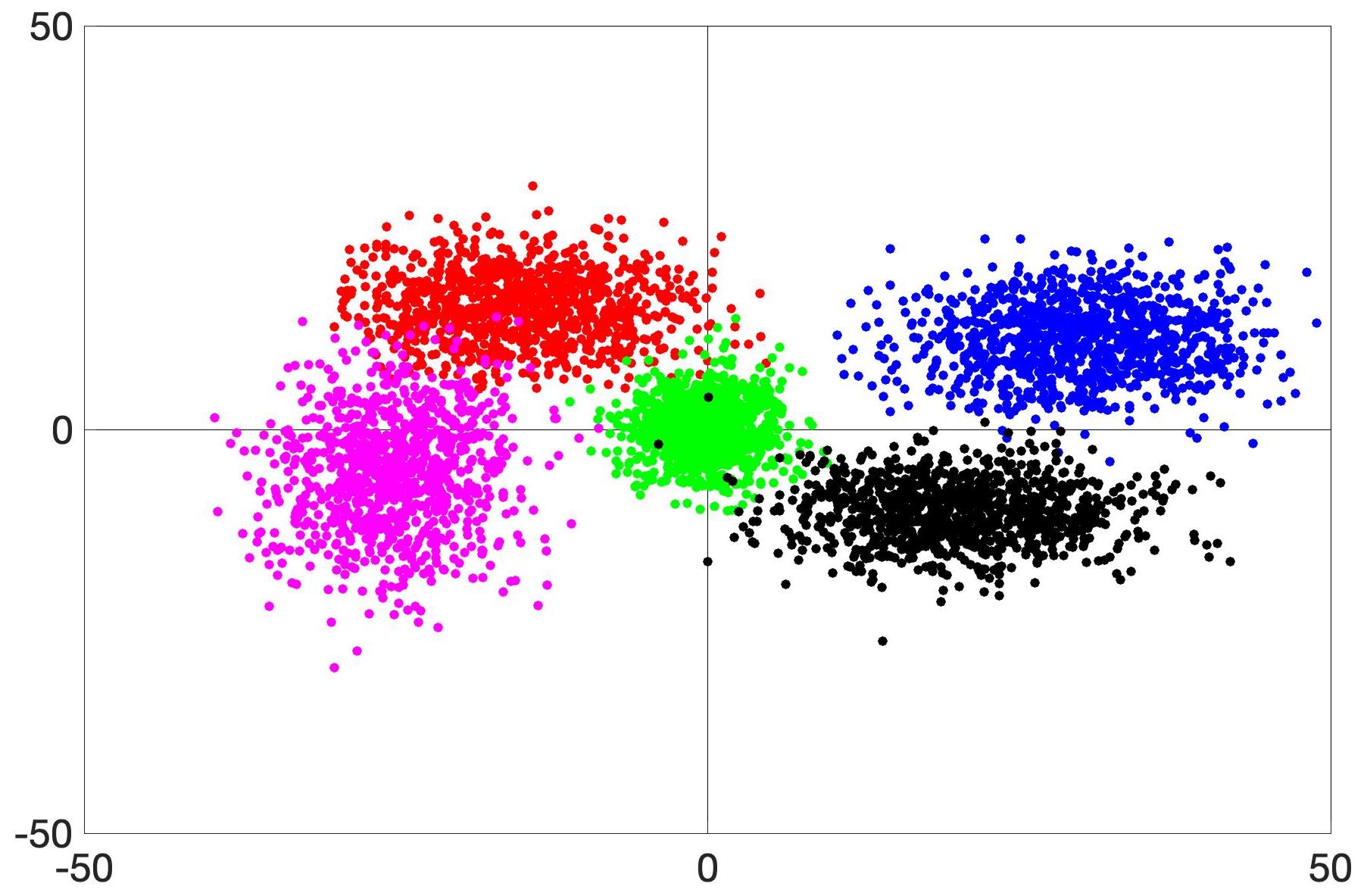} &
\includegraphics[height=2.3cm, width=3cm]{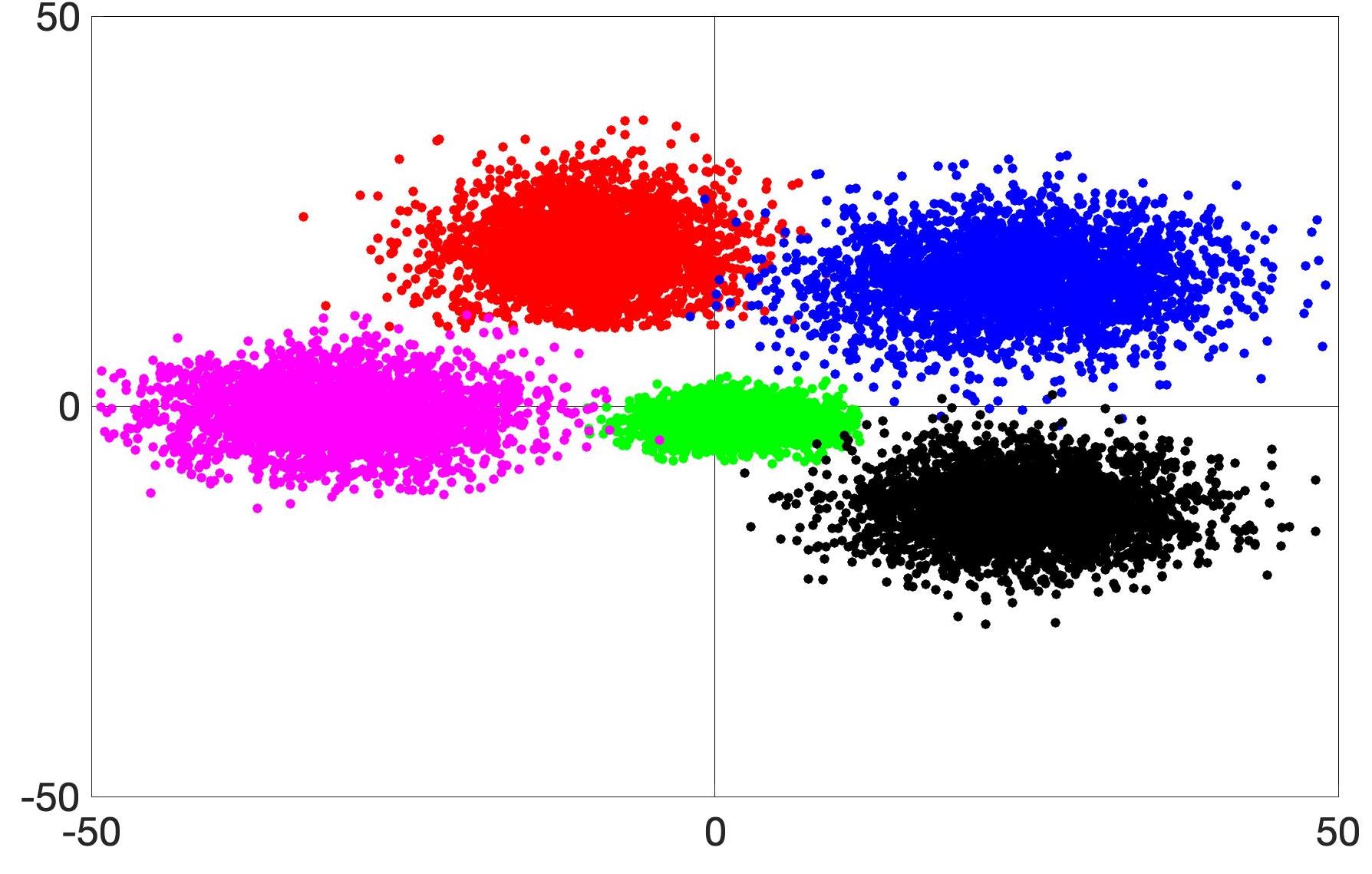} &
\includegraphics[height=2.3cm, width=3cm]{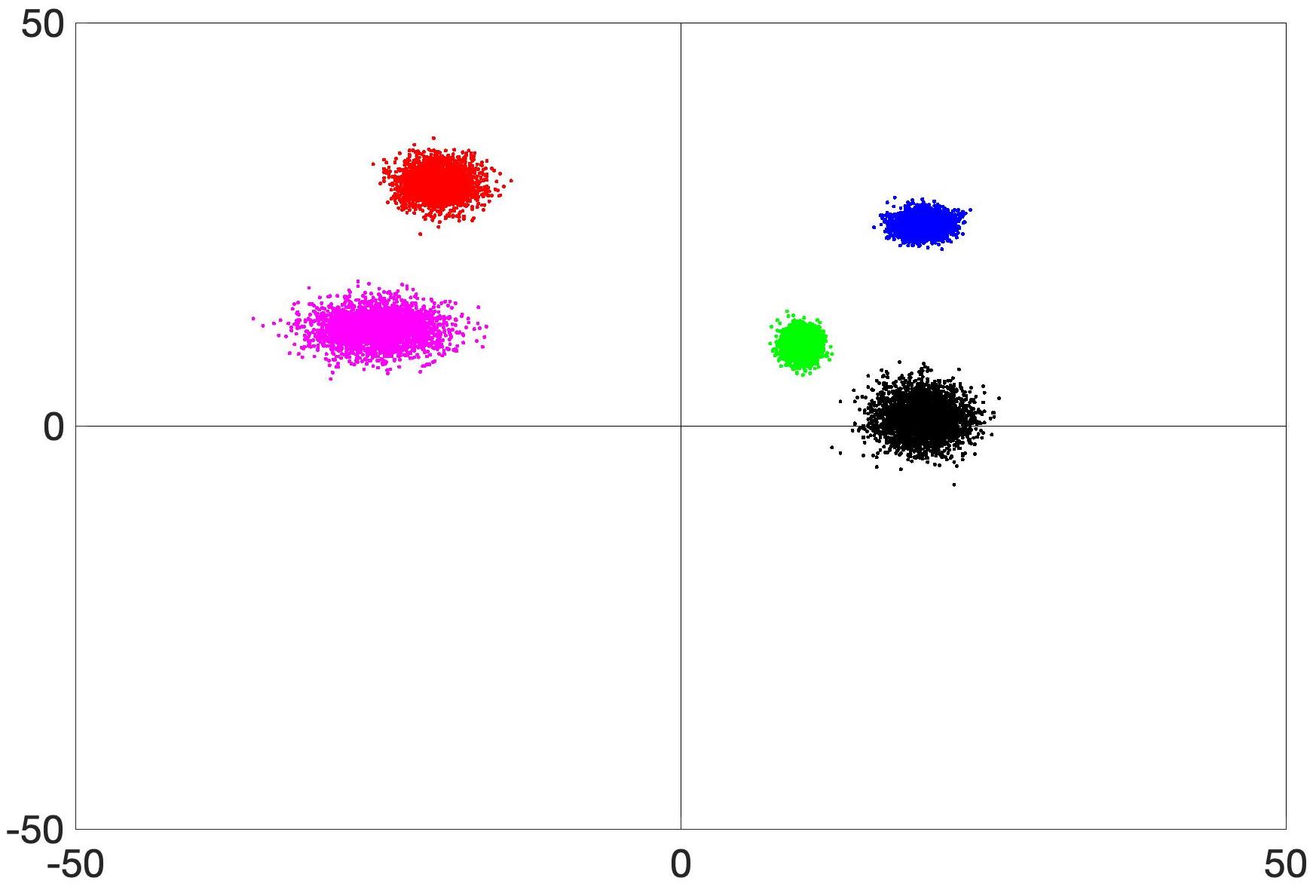} &
\includegraphics[height=2.3cm, width=3cm]{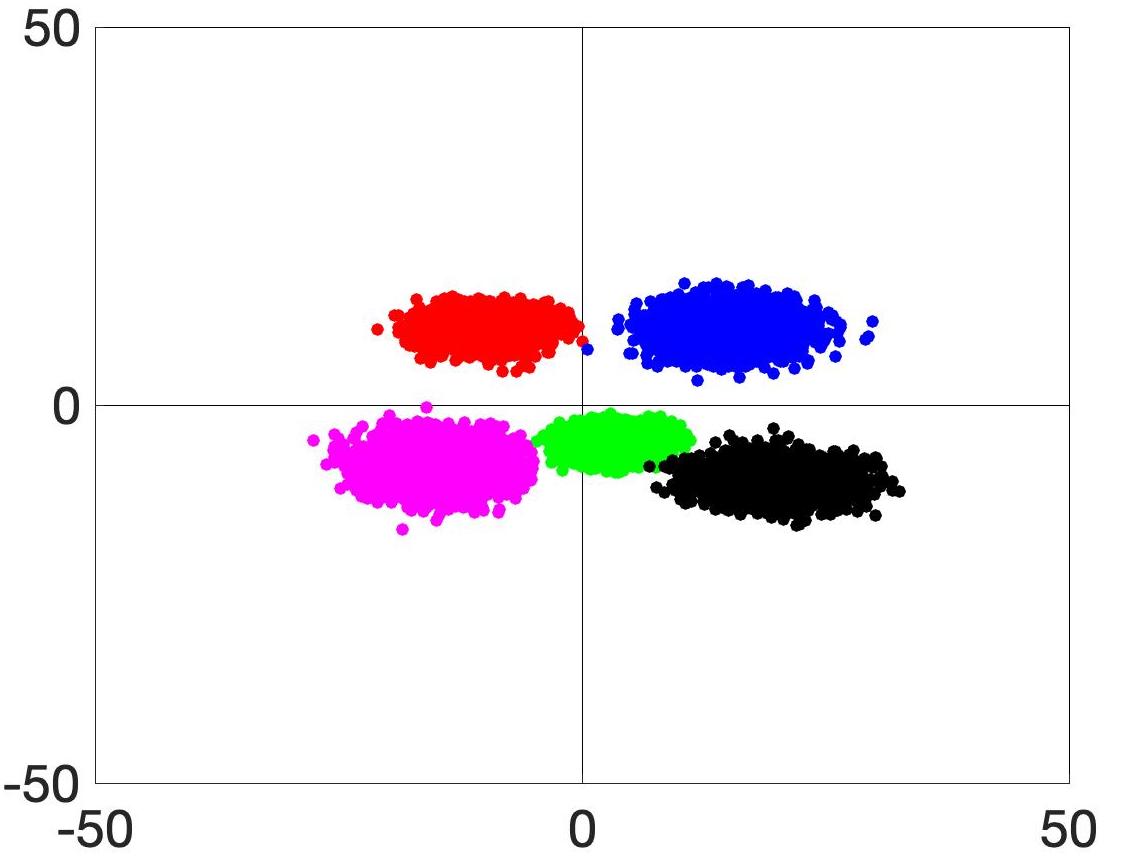} \\
(a) & (b) & (c) & (d) \\
\end{tabular}
\begin{tabular}{c}
\includegraphics[height=0.6cm, width=8cm]{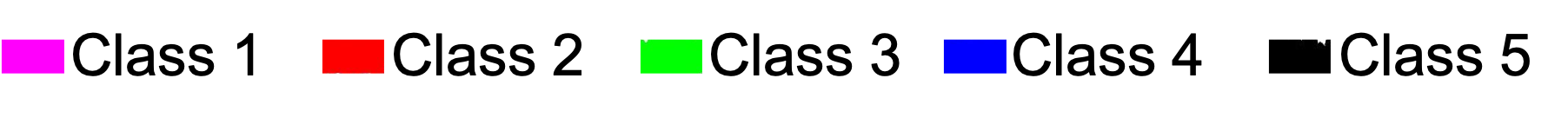}  \\
\end{tabular}
\caption{Feature visualizations: (a) Seen+Unseen classes from actual dataset; distribution of synthetic samples generated by b) $MM_{GZSL}$; (c) $MM_{GZSL}$ without self-supervision; (d) for \cite{WuCvpr20}. Different colours represent different classes. (b) is closer to (a), while (c) and (d) are quite different.}
\label{fig:FeatVis}
\end{figure}

{\subsubsection{Visualization of Synthetic Features}:}
%
Figure~\ref{fig:FeatVis} (a) shows t-SNE plot of real data features where the classes are spread over a wide area, with overlap amongst consecutive classes. Figures~\ref{fig:FeatVis} (b,c,d) show, respectively, distribution of synthetic features generated by MM$_{GZSL}$, MM$_{wCPC}$ and \cite{WuCvpr20}. MM$_{GZSL}$ features are the most similar to the original data. MM$_{wCPC}$ and \cite{WuCvpr20} synthesize sub-optimal feature representation of actual features, resulting in poor classification performance on unseen classes.

%

%% file: DGen_Concl.tex

\section{Conclusion}

We propose a GZSL approach without relying on class attribute vectors.  
Our novel method can accurately predict Gleason grades from MR images  with lower resolution and less information content than histopathology images. This has the potential to improve accuracy of early detection and staging of PCa. 
A self-supervised component ensures the semantic gap between Seen and Unseen classes is easily covered. The distribution of synthetic features generated by our method are close to the actual distribution, while removing the self-supervised term results in unrealistic distributions. 
Results show our method's superior performance and 
 synergy between different loss terms leads to improved GZSL classification.
We observe failure cases where the acquired MR images are not of sufficiently good resolution to allow accurate registration of histopathology and MRI. Future work will aim to address this issue.

%% file: ms.bbl
\begin{thebibliography}{100}
\providecommand{\url}[1]{\texttt{#1}}
\providecommand{\urlprefix}{URL }
\providecommand{\doi}[1]{https://doi.org/#1}

\bibitem{Pat15}
Antony, B., Sedai, S., Mahapatra, D., Garnavi, R.: Real-time passive monitoring
  and assessment of pediatric eye health. In: US Patent App. 16/178,757 (2020)

\bibitem{AMDIM}
Bachman, P., Hjelm, R.D., Buchwalter, W.: Learning representations by
  maximizing mutual information across views. In: In Proc. NeurIPS. pp.
  15509--15519 (2019)

\bibitem{Pat7}
Bastide, P., Kiral-Kornek, I., Mahapatra, D., Saha, S., Vishwanath, A.,
  Cavallar, S.V.: Machine learned optimizing of health activity for
  participants during meeting times. In: US Patent App. 15/426,634 (2018)

\bibitem{Pat5}
Bastide, P., Kiral-Kornek, I., Mahapatra, D., Saha, S., Vishwanath, A.,
  Cavallar, S.V.: Visual health maintenance and improvement. In: US Patent
  9,993,385 (2018)

\bibitem{Health_p}
Bastide, P., Kiral-Kornek, I., Mahapatra, D., Saha, S., Vishwanath, A.,
  Cavallar, S.V.: Crowdsourcing health improvements routes. In: US Patent App.
  15/611,519 (2019)

\bibitem{BhattaMiccai20}
Bhattacharya, I., Seetharaman, A., Shao, W., Sood, R., Kunder, C.A., et~al.:
  Corrsignet: Learning correlated prostate cancer signatures from radiology and
  pathology images for improved computer aided diagnosis. In: In Proc MICCAI.
  pp. 315--325 (2020)

\bibitem{Mahapatra_CVIU2019}
Bozorgtabar, B., Mahapatra, D., von Teng, H., Pollinger, A., Ebner, L., Thiran,
  J.P., Reyes, M.: Informative sample generation using class aware generative
  adversarial networks for classification of chest xrays. Computer Vision and
  Image Understanding  \textbf{184},  57--65 (2019)

\bibitem{CVIU_Ar}
Bozorgtabar, B., Mahapatra, D., von Teng, H., Pollinger, A., Ebner, L., Thiran,
  J.P., Reyes, M.: Informative sample generation using class aware generative
  adversarial networks for classification of chest xrays. In: arXiv preprint
  arXiv:1904.10781 (2019)

\bibitem{Behzad_PR2020}
Bozorgtabar, B., Mahapatra, D., Thiran, J.P.: Exprada: Adversarial domain
  adaptation for facial expression analysis. In Press Pattern Recognition
  \textbf{100},  15--28 (2020)

\bibitem{Behzad_MICCAI20}
Bozorgtabar, B., Mahapatra, D., Thiran, J.P., Shao, L.: {SALAD}:
  Self-supervised aggregation learning for anomaly detection on x-rays. In: In
  Proc. MICCAI. pp. 468--478 (2020)

\bibitem{Salad_AR}
Bozorgtabar, B., Mahapatra, D., Vray, G., Thiran, J.P.: Anomaly detection on
  x-rays using self-supervised aggregation learning. In: arXiv preprint
  arXiv:2010.09856 (2020)

\bibitem{Frontiers2020}
Bozorgtabar, B., Mahapatra, D., Zlobec, I., Rau, T., Thiran, J.: Computational
  pathology. Frontiers in Medicine  \textbf{7} (2020)

\bibitem{Bozorgtabar_ICCV19}
Bozorgtabar, B., Rad, M.S., Mahapatra, D., Thiran, J.P.: Syndemo: Synergistic
  deep feature alignment for joint learning of depth and ego-motion. In: In
  Proc. IEEE ICCV (2019)

\bibitem{Panda}
Bulten, W., Pinckaers, H., van Boven, H., Vink, R., de~Bel, T., van Ginneken,
  B., van~der Laak, J., van~de Kaa, C.H., Litjens, G.: Automated deep-learning
  system for gleason grading of prostate cancer using biopsies: a diagnostic
  study. Lancet Oncology  \textbf{21}(2),  233--241 (2020)

\bibitem{GGL26}
Campanella, G., Silva, V.M., Fuchs, T.J.: Terabyte-scale deep multiple instance
  learning for classification and localization in pathology. In: arXiv preprint
  arXiv:1805.06983 (2018)

\bibitem{FelixEccv18}
Felix, R., Kumar, V., Reid, I., Carneiro, G.: Multi-modal cycle-consistent
  generalized zero-shot learning. In: ECCV. pp. 21--37 (2018)

\bibitem{Kaggle}
Foundation, C.H.: Diabetic retinopathy detection. [online]. available:
  https://www.kaggle.com/c/ diabetic-retinopathy-detection

\bibitem{Pat14}
Garnavi, R., Mahapatra, D., Roy, P., Tennakoon, R.: System and method to teach
  and evaluate image grading performance using prior learned expert knowledge
  base. In: US Patent App. 10,657,838 (2020)

\bibitem{ZGe_MTA2019}
Ge, Z., Mahapatra, D., Chang, X., Chen, Z., Chi, L., Lu, H.: Improving
  multi-label chest x-ray disease diagnosis by exploiting disease and health
  labels dependencies. In press Multimedia Tools and Application pp. 1--14
  (2019)

\bibitem{Xr_Ar}
Ge, Z., Mahapatra, D., Sedai, S., Garnavi, R., Chakravorty, R.: Chest x-rays
  classification: A multi-label and fine-grained problem. In: arXiv preprint
  arXiv:1807.07247 (2018)

\bibitem{Gutmann2010}
Gutmann, M., Hyvarinen, A.: A new estimation principle for unnormalized
  statistical models. In: In Proc. AISTATS. pp. 297--304 (2010)

\bibitem{ResNet50}
He, K., Zhang, X., Ren, S., Sun, J.: Deep residual learning for image
  recognition. In: In Proc. CVPR. pp. 770--778 (2016)

\bibitem{Pat17}
Hoog, J.D., Mahapatra, D., Garnavi, R., Jalali, F.: Personalized monitoring of
  injury rehabilitation through mobile device imaging. In: US Patent App.
  16/589,046 (2021)

\bibitem{Verma11}
Hu, Z., Yang, Z., Liang, X., Salakhutdinov, R., Xing, E.P.: Toward controlled
  generation of text. In: In Proc. ICML. pp. 1587--1596 (2017)

\bibitem{HuangCVPR19}
Huang, H., Wang, C., Yu, P.S., Wang, C.D.: Generative dual adversarial network
  for generalized zero-shot learning. In: The IEEE Conference on Computer
  Vision and Pattern Recognition (CVPR). pp. 801--810 (June 2019)

\bibitem{Lie_AR2}
Ju, L., Wang, X., Wang, L., Liu, T., Zhao, X., Drummond, T., Mahapatra, D., Ge,
  Z.: Relational subsets knowledge distillation for long-tailed retinal
  diseases recognition. In: arXiv preprint arXiv:2104.11057 (2021)

\bibitem{Lie_AR}
Ju, L., Wang, X., Wang, L., Mahapatra, D., Zhao, X., Harandi, M., Drummond, T.,
  Liu, T., Ge, Z.: Improving medical image classification with label noise
  using dual-uncertainty estimation. In: arXiv preprint arXiv:2103.00528 (2020)

\bibitem{JuJbhi2020}
Ju, L., Wang, X., Zhao, X., Lu, H., Mahapatra, D., Bonnington, P., Ge, Z.:
  Synergic adversarial label learning for grading retinal diseases via
  knowledge distillation and multi-task learning. IEEE JBHI  \textbf{100},
  1--14 (2020)

\bibitem{LieMiccai21}
Ju, L., Wang, X., Zhao, X., Lu, H., Mahapatra, D., Ge, Z.: Relational subsets
  knowledge distillation for long-tailed retinal diseases recognition. In: In
  MICCAI 2021. pp. 1--11 (2021)

\bibitem{KeshariCvpr20}
Keshari, R., Singh, R., Vatsa, M.: Generalized zero-shot learning via
  over-complete distribution. In: The IEEE Conference on Computer Vision and
  Pattern Recognition (CVPR). pp. 13300--13308 (June 2020)

\bibitem{Adam}
Kingma, D.P., Ba, J.: Adam: A method for stochastic optimization. In: arXiv
  preprint arXiv:1412.6980, (2014)

\bibitem{SelfPath}
Koohbanani, N.A., Unnikrishnan, B., Khurram, S.A., Krishnaswamy, P., Rajpoot,
  N.: Self-path: Self-supervision for classification of pathology images with
  limited annotations. In: arXiv:2008.05571 (2020)

\bibitem{Kuanar_AR1}
Kuanar, S., Athitsos, V., Mahapatra, D., Rajan, A.: Multi-scale deep learning
  architecture for nucleus detection in renal cell carcinoma microscopy image.
  In: arXiv preprint arXiv:2104.13557 (2021)

\bibitem{Kuanar_ICIP19}
Kuanar, S., Athitsos, V., Mahapatra, D., Rao, K., Akhtar, Z., Dasgupta, D.: Low
  dose abdominal ct image reconstruction: An unsupervised learning based
  approach. In: In Proc. IEEE ICIP. pp. 1351--1355 (2019)

\bibitem{Kuanar_AR2}
Kuanar, S., Mahapatra, D., Athitsos, V., Rao, K.: Gated fusion network for sao
  filter and inter frame prediction in versatile video coding. In: arXiv
  preprint arXiv:2105.12229 (2021)

\bibitem{Haze_Ar}
Kuanar, S., Rao, K., Mahapatra, D., Bilas, M.: Night time haze and glow removal
  using deep dilated convolutional network. In: arXiv preprint arXiv:1902.00855
  (2019)

\bibitem{KuanarVC}
Kuanar, S., Mahapatra, D., Bilas, M., Rao, K.: Multi-path dilated convolution
  network for haze and glow removal in night time images. The Visual Computer
  pp. 1--14 (2021)

\bibitem{KuangAMM14}
Kuang, H., Guthier, B., Saini, M., Mahapatra, D., Saddik, A.E.: A real-time
  smart assistant for video surveillance through handheld devices. In: In Proc:
  ACM Intl. Conf. Multimedia. pp. 917--920 (2014)

\bibitem{LiTMI_2015}
Li, Z., Mahapatra, D., J.Tielbeek, Stoker, J., van Vliet, L., Vos, F.: Image
  registration based on autocorrelation of local structure. IEEE Trans. Med.
  Imaging  \textbf{35}(1),  63--75 (2016)

\bibitem{Promise12}
Litjens, G., Toth, R., de~Ven, W., Hoeks, C., et~al: Evaluation of prostate
  segmentation algorithms for mri: The {PROMISE12} challenge. Medical Image
  Analysis  \textbf{18}(2),  359--373 (2014)

\bibitem{Felix8}
Long, Y., Liu, L., Shen, F., Shao, L., Li, X.: Zero-shot learning using
  synthesised unseen visual data with diffusion regularisation. IEEE Trans.
  Pattern Analysis Machine Intelligence  \textbf{40}(10),  2498 -- 2512 (2017)

\bibitem{LuMahmood}
Lu, M.Y., Chen, R.J., Wang, J., Dillon, D., Mahmood, F.: Semi-supervised
  histology classification using deep multiple instance learning and
  contrastive predictive coding. In: arXiv:1910.10825 (2019)

\bibitem{Felix34}
Maas, A.L., Hannun, A.Y., Ng, A.Y.: Rectifier nonlinearities improve neural
  network acoustic models. In: In Proc. ICML (2013)

\bibitem{MahapatraRegBook}
Mahapatra, D.: Elastic registration of cardiac perfusion images using saliency
  information. Sequence and Genome Analysis – Methods and Applications pp.
  351--364 (2011)

\bibitem{MahapatraMiccaiIAHBD11}
Mahapatra, D.: Neonatal brain mri skull stripping using graph cuts and shape
  priors. In: In Proc: MICCAI workshop on Image Analysis of Human Brain
  Development (IAHBD) (2011)

\bibitem{MahapatraMLMI12}
Mahapatra, D.: Cardiac lv and rv segmentation using mutual context information.
  In: Proc. MICCAI-MLMI. pp. 201--209 (2012)

\bibitem{MahapatraGRSPIE12}
Mahapatra, D.: Groupwise registration of dynamic cardiac perfusion images using
  temporal information and segmentation information. In: In Proc: SPIE Medical
  Imaging (2012)

\bibitem{MahapatraSTACOM12}
Mahapatra, D.: Landmark detection in cardiac mri using learned local image
  statistics. In: Proc. MICCAI-Statistical Atlases and Computational Models of
  the Heart. Imaging and Modelling Challenges (STACOM). pp. 115--124 (2012)

\bibitem{MahapatraJDISkull2012}
Mahapatra, D.: Skull stripping of neonatal brain mri: Using prior shape
  information with graphcuts. J. Digit. Imaging  \textbf{25}(6),  802--814
  (2012)

\bibitem{MahapatraJDIGCSP2013}
Mahapatra, D.: Cardiac image segmentation from cine cardiac mri using graph
  cuts and shape priors. J. Digit. Imaging  \textbf{26}(4),  721--730 (2013)

\bibitem{MahapatraJDIMutCont2013}
Mahapatra, D.: Cardiac mri segmentation using mutual context information from
  left and right ventricle. J. Digit. Imaging  \textbf{26}(5),  898--908 (2013)

\bibitem{MahapatraProISBI13}
Mahapatra, D.: Graph cut based automatic prostate segmentation using learned
  semantic information. In: Proc. IEEE ISBI. pp. 1304--1307 (2013)

\bibitem{MahapatraJDIJSGR2013}
Mahapatra, D.: Joint segmentation and groupwise registration of cardiac
  perfusion images using temporal information. J. Digit. Imaging
  \textbf{26}(2),  173--182 (2013)

\bibitem{MahapatraJDI_Cardiac_FSL}
Mahapatra, D.: An automated approach to cardiac rv segmentation from mri using
  learned semantic information and graph cuts. J. Digit. Imaging.
  \textbf{27}(6),  794--804 (2014)

\bibitem{Mahapatra_LME_CVIU}
Mahapatra, D.: Combining multiple expert annotations using semi-supervised
  learning and graph cuts for medical image segmentation. Computer Vision and
  Image Understanding  \textbf{151}(1),  114--123 (2016)

\bibitem{Mahapatra_OMIA16}
Mahapatra, D.: Retinal image quality classification using neurobiological
  models of the human visual system. In: In Proc. MICCAI-OMIA. pp.~1--8 (2016)

\bibitem{LME_Ar}
Mahapatra, D.: Consensus based medical image segmentation using semi-supervised
  learning and graph cuts. In: arXiv preprint arXiv:1612.02166 (2017)

\bibitem{Mahapatra_LME_PR2017}
Mahapatra, D.: Semi-supervised learning and graph cuts for consensus based
  medical image segmentation. Pattern Recognition  \textbf{63}(1),  700--709
  (2017)

\bibitem{AMD_OCT}
Mahapatra, D.: Amd severity prediction and explainability using image
  registration and deep embedded clustering. In: arXiv preprint
  arXiv:1907.03075 (2019)

\bibitem{GANReg2_Ar}
Mahapatra, D.: Generative adversarial networks and domain adaptation for
  training data independent image registration. In: arXiv preprint
  arXiv:1910.08593 (2019)

\bibitem{DART2020_Ar}
Mahapatra, D.: Registration of histopathogy images using structural information
  from fine grained feature maps. In: arXiv preprint arXiv:2007.02078 (2020)

\bibitem{TMI2021_Ar}
Mahapatra, D.: Interpretability-driven sample selection using self supervised
  learning for disease classification and segmentation. In: arXiv preprint
  arXiv:2104.06087 (2021)

\bibitem{DARTSyn_Ar}
Mahapatra, D.: Learning of inter-label geometric relationships using
  self-supervised learning: Application to gleason grade segmentation. In:
  arXiv preprint arXiv:2110.00404 (2021)

\bibitem{Misc}
Mahapatra, D., Agarwal, K., Khosrowabadi, R., Prasad, D.: Recent advances in
  statistical data and signal analysis: Application to real world diagnostics
  from medical and biological signals. In: Computational and mathematical
  methods in medicine (2016)

\bibitem{MahapatraGAN_ISBI18}
Mahapatra, D., Antony, B., Sedai, S., Garnavi, R.: Deformable medical image
  registration using generative adversarial networks. In: In Proc. IEEE ISBI.
  pp. 1449--1453 (2018)

\bibitem{ISR_Ar}
Mahapatra, D., Bozorgtabar, B.: Retinal vasculature segmentation using local
  saliency maps and generative adversarial networks for image super resolution.
  In: arXiv preprint arXiv:1710.04783 (2017)

\bibitem{PGAN_Ar}
Mahapatra, D., Bozorgtabar, B.: Progressive generative adversarial networks for
  medical image super resolution. In: arXiv preprint arXiv:1902.02144 (2019)

\bibitem{Mahapatra_CMIG2019}
Mahapatra, D., Bozorgtabar, B., Garnavi, R.: Image super-resolution using
  progressive generative adversarial networks for medical image analysis.
  Computerized Medical Imaging and Graphics  \textbf{71},  30--39 (2019)

\bibitem{Mahapatra_CVAMD2021}
Mahapatra, D., Bozorgtabar, B., Ge, Z.: Medical image classification using
  generalized zero shot learning. In: In IEEE CVAMD 2021. pp. 3344--3353 (2021)

\bibitem{Mahapatra_DART21a}
Mahapatra, D., Bozorgtabar, B., Kuanar, S., Ge, Z.: Self-supervised multimodal
  generalized zero shot learning for gleason grading. In: In MICCAI-DART 2021.
  pp. 1--11 (2021)

\bibitem{Mahapatra_CVPR2020}
Mahapatra, D., Bozorgtabar, B., Shao, L.: Pathological retinal region
  segmentation from oct images using geometric relation based augmentation. In:
  In Proc. IEEE CVPR. pp. 9611--9620 (2020)

\bibitem{CVPR2020_Ar}
Mahapatra, D., Bozorgtabar, B., Thiran, J.P., Shao, L.: Pathological retinal
  region segmentation from oct images using geometric relation based
  augmentation. In: arXiv preprint arXiv:2003.14119 (2020)

\bibitem{Mahapatra_MICCAI20}
Mahapatra, D., Bozorgtabar, B., Thiran, J.P., Shao, L.: Structure preserving
  stain normalization of histopathology images using self supervised semantic
  guidance. In: In Proc. MICCAI. pp. 309--319 (2020)

\bibitem{Stain_AR}
Mahapatra, D., Bozorgtabar, B., Thiran, J.P., Shao, L.: Structure preserving
  stain normalization of histopathology images using self supervised semantic
  guidance. In: arXiv preprint arXiv:2008.02101 (2020)

\bibitem{Mahapatra_MICCAI17}
Mahapatra, D., Bozorgtabar, S., Hewavitahranage, S., Garnavi, R.: Image super
  resolution using generative adversarial networks and local saliencymaps for
  retinal image analysis,. In: In Proc. MICCAI. pp. 382--390 (2017)

\bibitem{MahapatraAL_MICCAI18}
Mahapatra, D., Bozorgtabar, S., Thiran, J.P., Reyes, M.: Efficient active
  learning for image classification and segmentation using a sample selection
  and conditional generative adversarial network. In: In Proc. MICCAI (2). pp.
  580--588 (2018)

\bibitem{Mahapatra_OMIA15}
Mahapatra, D., Buhmann, J.: Obtaining consensus annotations for retinal image
  segmentation using random forest and graph cuts. In: In Proc. OMIA. pp.
  41--48 (2015)

\bibitem{Mahapatra_MLMI15_Prostate}
Mahapatra, D., Buhmann, J.: Visual saliency based active learning for prostate
  mri segmentation. In: In Proc. MLMI. pp. 9--16 (2015)

\bibitem{Mahapatra_SSLAL_Pro_JMI}
Mahapatra, D., Buhmann, J.: Visual saliency based active learning for prostate
  mri segmentation. SPIE Journal of Medical Imaging  \textbf{3}(1) (2016)

\bibitem{MahapatraRVISBI13}
Mahapatra, D., Buhmann, J.: Automatic cardiac rv segmentation using semantic
  information with graph cuts. In: Proc. IEEE ISBI. pp. 1094--1097 (2013)

\bibitem{MahapatraTIP_RF2014}
Mahapatra, D., Buhmann, J.: Analyzing training information from random forests
  for improved image segmentation. IEEE Trans. Imag. Proc.  \textbf{23}(4),
  1504--1512 (2014)

\bibitem{MahapatraTBME_Pro2014}
Mahapatra, D., Buhmann, J.: Prostate mri segmentation using learned semantic
  knowledge and graph cuts. IEEE Trans. Biomed. Engg.  \textbf{61}(3),
  756--764 (2014)

\bibitem{MahapatraISBI15_Optic}
Mahapatra, D., Buhmann, J.: A field of experts model for optic cup and disc
  segmentation from retinal fundus images. In: In Proc. IEEE ISBI. pp. 218--221
  (2015)

\bibitem{Pat2}
Mahapatra, D., Garnavi, R., Roy, P., Tennakoon, R.: System and method to teach
  and evaluate image grading performance using prior learned expert knowledge
  base. In: US Patent App. 15/459,457 (2018)

\bibitem{Pat3}
Mahapatra, D., Garnavi, R., Roy, P., Tennakoon, R.: System and method to teach
  and evaluate image grading performance using prior learned expert knowledge
  base. In: US Patent App. 15/814,590 (2018)

\bibitem{Pat11}
Mahapatra, D., Garnavi, R., Sedai, S., Roy, P.: Joint segmentation and
  characteristics estimation in medical images. In: US Patent App. 15/234,426
  (2017)

\bibitem{Pat10}
Mahapatra, D., Garnavi, R., Sedai, S., Roy, P.: Retinal image quality
  assessment, error identification and automatic quality correction. In: US
  Patent 9,779,492 (2017)

\bibitem{Pat6}
Mahapatra, D., Garnavi, R., Sedai, S., Tennakoon, R.: Classification of
  severity of pathological condition using hybrid image representation. In: US
  Patent App. 15/426,634 (2018)

\bibitem{Pat4}
Mahapatra, D., Garnavi, R., Sedai, S., Tennakoon, R.: Generating an enriched
  knowledge base from annotated images. In: US Patent App. 15/429,735 (2018)

\bibitem{Pat8}
Mahapatra, D., Garnavi, R., Sedai, S., Tennakoon, R., Chakravorty, R.: Early
  prediction of age related macular degeneration by image reconstruction. In:
  US Patent App. 15/854,984 (2018)

\bibitem{Pat9}
Mahapatra, D., Garnavi, R., Sedai, S., Tennakoon, R., Chakravorty, R.: Early
  prediction of age related macular degeneration by image reconstruction. In:
  US Patent 9,943,225 (2018)

\bibitem{GANReg1_Ar}
Mahapatra, D., Ge, Z.: Combining transfer learning and segmentation information
  with gans for training data independent image registration. In: arXiv
  preprint arXiv:1903.10139 (2019)

\bibitem{Mahapatra_ISBI19}
Mahapatra, D., Ge, Z.: Training data independent image registration with gans
  using transfer learning and segmentation information. In: In Proc. IEEE ISBI.
  pp. 709--713 (2019)

\bibitem{Mahapatra_PR2020}
Mahapatra, D., Ge, Z.: Training data independent image registration using
  generative adversarial networks and domain adaptation. Pattern Recognition
  \textbf{100},  1--14 (2020)

\bibitem{Pat13}
Mahapatra, D., Ge, Z., Sedai, S.: Joint registration and segmentation of images
  using deep learning. In: US Patent App. 16/001,566 (2019)

\bibitem{Mahapatra_MLMI18}
Mahapatra, D., Ge, Z., Sedai, S., Chakravorty., R.: Joint registration and
  segmentation of xray images using generative adversarial networks. In: In
  Proc. MICCAI-MLMI. pp. 73--80 (2018)

\bibitem{Mahapatra_JSTSP2014}
Mahapatra, D., Gilani, S., Saini., M.: Coherency based spatio-temporal saliency
  detection for video object segmentation. IEEE Journal of Selected Topics in
  Signal Processing.  \textbf{8}(3),  454--462 (2014)

\bibitem{MahapatraTMI_CD2013}
Mahapatra, D., J.Tielbeek, Makanyanga, J., Stoker, J., Taylor, S., Vos, F.,
  Buhmann, J.: Automatic detection and segmentation of crohn's disease tissues
  from abdominal mri. IEEE Trans. Med. Imaging  \textbf{32}(12),  1232--1248
  (2013)

\bibitem{MahapatraISBI_CD2014}
Mahapatra, D., J.Tielbeek, Makanyanga, J., Stoker, J., Taylor, S., Vos, F.,
  Buhmann, J.: Active learning based segmentation of crohn's disease using
  principles of visual saliency. In: Proc. IEEE ISBI. pp. 226--229 (2014)

\bibitem{Mahapatra_ABD2014}
Mahapatra, D., J.Tielbeek, Makanyanga, J., Stoker, J., Taylor, S., Vos, F.,
  Buhmann, J.: Combining multiple expert annotations using semi-supervised
  learning and graph cuts for crohn's disease segmentation. In: In Proc:
  MICCAI-ABD (2014)

\bibitem{MahapatraJDICD2013}
Mahapatra, D., J.Tielbeek, Vos, F., Buhmann, J.: A supervised learning approach
  for crohn's disease detection using higher order image statistics and a novel
  shape asymmetry measure. J. Digit. Imaging  \textbf{26}(5),  920--931 (2013)

\bibitem{Mahapatra_DART21b}
Mahapatra, D., Kuanar, S., Bozorgtabar, B., Ge, Z.: Self-supervised learning of
  inter-label geometric relationships for gleason grade segmentation. In: In
  MICCAI-DART 2021. pp. 1--11 (2021)

\bibitem{MahapatraISBI15_JSGR}
Mahapatra, D., Li, Z., Vos, F., Buhmann, J.: Joint segmentation and groupwise
  registration of cardiac dce mri using sparse data representations. In: In
  Proc. IEEE ISBI. pp. 1312--1315 (2015)

\bibitem{MahapatraICIT06}
Mahapatra, D., Routray, A., Mishra, C.: An active snake model for
  classification of extreme emotions. In: IEEE International Conference on
  Industrial Technology (ICIT). pp. 2195--2199 (2006)

\bibitem{Mahapatra_EMBC16}
Mahapatra, D., Roy, P., Sedai, S., Garnavi, R.: A cnn based neurobiology
  inspired approach for retinal image quality assessment. In: In Proc. EMBC.
  pp. 1304--1307 (2016)

\bibitem{Mahapatra_MLMI16}
Mahapatra, D., Roy, P., Sedai, S., Garnavi, R.: Retinal image quality
  classification using saliency maps and cnns. In: In Proc. MICCAI-MLMI. pp.
  172--179 (2016)

\bibitem{MahapatraICBME08_Retrieve}
Mahapatra, D., Roy, S., Sun, Y.: Retrieval of mr kidney images by incorporating
  shape information in histogram of low level features. In: In 13th
  International Conference on Biomedical Engineering. pp. 661--664 (2009)

\bibitem{Pat12}
Mahapatra, D., Saha, S., Vishwanath, A., Bastide, P.: Generating hyperspectral
  image database by machine learning and mapping of color images to
  hyperspectral domain. In: US Patent App. 15/949,528 (2019)

\bibitem{MahapatraTrack_Book}
Mahapatra, D., Saini, M.: A particle filter framework for object tracking using
  visual-saliency information. Intelligent Multimedia Surveillance pp. 133--147
  (2013)

\bibitem{MahapatraICME08}
Mahapatra, D., Saini, M., Sun, Y.: Illumination invariant tracking in office
  environments using neurobiology-saliency based particle filter. In: IEEE
  ICME. pp. 953--956 (2008)

\bibitem{MahapatraMICCAI_CD2013}
Mahapatra, D., Sch$\ddot{u}$ffler, P., Tielbeek, J., Vos, F., Buhmann, J.:
  Semi-supervised and active learning for automatic segmentation of crohn's
  disease. In: Proc. MICCAI, Part 2. pp. 214--221 (2013)

\bibitem{RegGan_Ar}
Mahapatra, D., Sedai, S., Garnavi, R.: Elastic registration of medical images
  with gans. In: arXiv preprint arXiv:1805.02369 (2018)

\bibitem{Pat16}
Mahapatra, D., Sedai, S., Halupka, K.: Uncertainty region based image
  enhancement. In: US Patent App. 10,832,074 (2020)

\bibitem{Covi19_Ar}
Mahapatra, D., Singh, A.: Ct image synthesis using weakly supervised
  segmentation and geometric inter-label relations for covid image analysis.
  In: arXiv preprint arXiv:2106.10230 (2021)

\bibitem{MahapatraMiccai08}
Mahapatra, D., Sun, Y.: Nonrigid registration of dynamic renal {MR} images
  using a saliency based {MRF} model. In: Proc. MICCAI. pp. 771--779 (2008)

\bibitem{MahapatraISBI08}
Mahapatra, D., Sun, Y.: Registration of dynamic renal mr images using
  neurobiological model of saliency. In: Proc. ISBI. pp. 1119--1122 (2008)

\bibitem{MahapatraICBME08_Sal}
Mahapatra, D., Sun, Y.: Using saliency features for graphcut segmentation of
  perfusion kidney images. In: In 13th International Conference on Biomedical
  Engineering (2008)

\bibitem{MahapatraMiccai10}
Mahapatra, D., Sun, Y.: Joint registration and segmentation of dynamic cardiac
  perfusion images using mrfs. In: Proc. MICCAI. pp. 493--501 (2010)

\bibitem{MahapatraICIP10}
Mahapatra, D., Sun., Y.: An mrf framework for joint registration and
  segmentation of natural and perfusion images. In: Proc. IEEE ICIP. pp.
  1709--1712 (2010)

\bibitem{MahapatraICDIP10a}
Mahapatra, D., Sun, Y.: Retrieval of perfusion images using cosegmentation and
  shape context information. In: Proc. APSIPA Annual Summit and Conference
  (ASC) (2010)

\bibitem{MahapatraEURASIP2010}
Mahapatra, D., Sun, Y.: Rigid registration of renal perfusion images using a
  neurobiology based visual saliency model. EURASIP Journal on Image and Video
  Processing. pp. 1--16 (2010)

\bibitem{MahapatraICDIP10b}
Mahapatra, D., Sun, Y.: A saliency based mrf method for the joint registration
  and segmentation of dynamic renal mr images. In: Proc. ICDIP (2010)

\bibitem{MahapatraTBME2011}
Mahapatra, D., Sun, Y.: Mrf based intensity invariant elastic registration of
  cardiac perfusion images using saliency information. IEEE Trans. Biomed.
  Engg.  \textbf{58}(4),  991--1000 (2011)

\bibitem{MahapatraMiccai11}
Mahapatra, D., Sun, Y.: Orientation histograms as shape priors for left
  ventricle segmentation using graph cuts. In: In Proc: MICCAI. pp. 420--427
  (2011)

\bibitem{MahapatraTIP2012}
Mahapatra, D., Sun, Y.: Integrating segmentation information for improved
  mrf-based elastic image registration. IEEE Trans. Imag. Proc.
  \textbf{21}(1),  170--183 (2012)

\bibitem{MahapatraABD12}
Mahapatra, D., Tielbeek, J., Buhmann, J., Vos, F.: A supervised learning based
  approach to detect crohn's disease in abdominal mr volumes. In: Proc. MICCAI
  workshop Computational and Clinical Applications in Abdominal
  Imaging(MICCAI-ABD). pp. 97--106 (2012)

\bibitem{MahapatraCDFssISBI13}
Mahapatra, D., Tielbeek, J., Vos, F., ., J.B.: Crohn's disease tissue
  segmentation from abdominal mri using semantic information and graph cuts.
  In: Proc. IEEE ISBI. pp. 358--361 (2013)

\bibitem{MahapatraCDSPIE13}
Mahapatra, D., Tielbeek, J., Vos, F., Buhmann, J.: Localizing and segmenting
  crohn's disease affected regions in abdominal mri using novel context
  features. In: Proc. SPIE Medical Imaging (2013)

\bibitem{MahapatraWssISBI13}
Mahapatra, D., Tielbeek, J., Vos, F., Buhmann, J.: Weakly supervised semantic
  segmentation of crohn's disease tissues from abdominal mri. In: Proc. IEEE
  ISBI. pp. 832--835 (2013)

\bibitem{MahapatraISBI15_CD}
Mahapatra, D., Vos, F., Buhmann, J.: Crohn's disease segmentation from mri
  using learned image priors. In: In Proc. IEEE ISBI. pp. 625--628 (2015)

\bibitem{Mahapatra_SSLAL_CD_CMPB}
Mahapatra, D., Vos, F., Buhmann, J.: Active learning based segmentation of
  crohns disease from abdominal mri. Computer Methods and Programs in
  Biomedicine  \textbf{128}(1),  75--85 (2016)

\bibitem{MahapatraSPIE08}
Mahapatra, D., Winkler, S., Yen, S.: Motion saliency outweighs other low-level
  features while watching videos. In: SPIE HVEI. pp. 1--10 (2008)

\bibitem{Mahapatra_Thesis}
Mahapatra, D.: Registration and segmentation methodology for perfusion mr
  images: Application to cardiac and renal images. - pp.~-- (2011)

\bibitem{MahapatraTh2012}
Mahapatra, D.: Registration and segmentation methodology for perfusion mr
  images: Application to cardiac and renal images. - pp.~-- (2011)

\bibitem{MahapatraTMI2021}
Mahapatra, D., Poellinger, A., Shao, L., Reyes, M.: Interpretability-driven
  sample selection using self supervised learning for disease classification
  and segmentation. IEEE TMI pp. 1--15 (2021)

\bibitem{MinCVPR20}
Min, S., Yao, H., Xie, H., Wang, C., Zha, Z.J., Zhang, Y.: Domain-aware visual
  bias eliminating for generalized zero-shot learning. In: The IEEE Conference
  on Computer Vision and Pattern Recognition (CVPR). pp. 12664--12673 (June
  2020)

\bibitem{Mnih2013}
Mnih, A., Kavukcuoglu, K.: Learning word embeddings efficiently with
  noise-contrastive estimation. In: In Proc. NeurIPS. p. 2265–2273 (2013)

\bibitem{Felix35}
Nair, V., Hinton, G.E.: Rectified linear units improve restricted boltzmann
  machines. In: In Proc. ICML. pp. 807--814 (2010)

\bibitem{Bhatta12}
Nyul, L., Udupa, J., Zhang, X.: New variants of a method of mri scale
  standardization. IEEE Trans. Medical Imaging  \textbf{19}(2),  143--150
  (2000)

\bibitem{Oord}
van~den Oord, A., Li, Y., Vinyals, O.: Representation learning with contrastive
  predictive coding. In: arXiv:1807.03748 (2018)

\bibitem{PandeyiMIMIC2021}
Pandey, A., Paliwal, B., Dhall, A., Subramanian, R., Mahapatra, D.: This
  explains that: Congruent image--report generation for explainable medical
  image analysis with cyclic generative adversarial networks. In: In
  MICCAI-iMIMIC 2021. pp. 1--11 (2021)

\bibitem{Roy_DICTA16}
Roy, P., Chakravorty, R., Sedai, S., Mahapatra, D., Garnavi, R.: Automatic eye
  type detection in retinal fundus image using fusion of transfer learning and
  anatomical features. In: In Proc. DICTA. pp.~1--7 (2016)

\bibitem{Roy_ISBI17}
Roy, P., Tennakoon, R., Cao, K., Sedai, S., Mahapatra, D., Maetschke, S.,
  Garnavi, R.: A novel hybrid approach for severity assessment of diabetic
  retinopathy in colour fundus images,. In: In Proc. IEEE ISBI. pp. 1078--1082
  (2017)

\bibitem{Pat18}
Roy, P., Mahapatra, D., Garnavi, R., Tennakoon, R.: System and method to teach
  and evaluate image grading performance using prior learned expert knowledge
  base. In: US Patent App. 10,984,674 (2021)

\bibitem{Bhatta11}
Rusu, M., Shao, W., Kunder, C.A., Wang, J.B., Soerensen, S.J.C., et~al.:
  Registration of pre-surgical mri and histopathology images from radical
  prostatectomy via rapsodi. Medical Physics  \textbf{47}(9),  4177--4188
  (2020)

\bibitem{sZoom_Ar}
Saini, M., Guthier, B., Kuang, H., Mahapatra, D., Saddik, A.: szoom: A
  framework for automatic zoom into high resolution surveillance videos. In:
  arXiv preprint arXiv:1909.10164 (2019)

\bibitem{Schuffler_ABD2013}
Sch$\ddot{u}$ffler, P., Mahapatra, D., Tielbeek, J., Vos, F., Makanyanga, J.,
  Pends, D., Nio, C., Stoker, J., Taylor, S., Buhmann, J.: A model development
  pipeline for crohns disease severity assessment from magnetic resonance
  images. In: In Proc: MICCAI-ABD (2013)

\bibitem{Schuffler_ABD2014}
Sch$\ddot{u}$ffler, P., Mahapatra, D., Tielbeek, J., Vos, F., Makanyanga, J.,
  Pends, D., Nio, C., Stoker, J., Taylor, S., Buhmann, J.: Semi automatic
  crohns disease severity assessment on mr imaging. In: In Proc: MICCAI-ABD
  (2014)

\bibitem{Schuffler_ABD2014_2}
Schuffler, P.J., Mahapatra, D., Vos, F., Buhmann, J.: Computer aided crohn’s
  disease severity assessment in mri. In: VIGOR++ Workshop 2014-Showcase of
  Research Outcomes and Future Outlook. pp.~-- (2014)

\bibitem{Sedai_OMIA18}
Sedai, S., Mahapatra, D., Antony, B., Garnavi, R.: Joint segmentation and
  uncertainty visualization of retinal layers in optical coherence tomography
  images using bayesian deep learning. In: In Proc. MICCAI-OMIA. pp. 219--227
  (2018)

\bibitem{Sedai_MLMI18}
Sedai, S., Mahapatra, D., Ge, Z., Chakravorty, R., Garnavi, R.: Deep multiscale
  convolutional feature learning for weakly supervised localization of chest
  pathologies in x-ray images. In: In Proc. MICCAI-MLMI. pp. 267--275 (2018)

\bibitem{Sedai_MICCAI17}
Sedai, S., Mahapatra, D., Hewavitharanage, S., Maetschke, S., Garnavi, R.:
  Semi-supervised segmentation of optic cup in retinal fundus images using
  variational autoencoder,. In: In Proc. MICCAI. pp. 75--82 (2017)

\bibitem{Sedai_EMBC16}
Sedai, S., Roy, P., Mahapatra, D., Garnavi, R.: Segmentation of optic disc and
  optic cup in retinal fundus images using shape regression. In: In Proc. EMBC.
  pp. 3260--3264 (2016)

\bibitem{Sedai_OMIA16}
Sedai, S., Roy, P., Mahapatra, D., Garnavi, R.: Segmentation of optic disc and
  optic cup in retinal images using coupled shape regression. In: In Proc.
  MICCAI-OMIA. pp.~1--8 (2016)

\bibitem{SrivastavaFAIR2021}
Srivastava, S., Yaqub, M., Nandakumar, K., Ge, Z., Mahapatra, D.: Continual
  domain incremental learning for chest x-ray classification in low-resource
  clinical settings. In: In MICCAI-FAIR 2021. pp. 1--11 (2021)

\bibitem{Tennakoon_OMIA16}
Tennakoon, R., Mahapatra, D., Roy, P., Sedai, S., Garnavi, R.: Image quality
  classification for dr screening using convolutional neural networks. In: In
  Proc. MICCAI-OMIA. pp. 113--120 (2016)

\bibitem{TongDART20}
Tong, J., Mahapatra, D., Bonnington, P., Drummond, T., Ge, Z.: Registration of
  histopathology images using self supervised fine grained feature maps. In: In
  Proc. MICCAI-DART Workshop. pp. 41--51 (2020)

\bibitem{MonusacTMI}
Verma, R., Kumar, N., Patil, A., et. al.: Monusac2020: A multi-organ nuclei
  segmentation and classification challenge. IEEE TMI pp. 1--14 (2021)

\bibitem{VosEMBC}
Vos, F.M., Tielbeek, J., Naziroglu, R., Li, Z., Sch$\ddot{u}$ffler, P.,
  Mahapatra, D., Wiebel, A., Lavini, C., Buhmann, J., Hege, H., Stoker, J., van
  Vliet, L.: Computational modeling for assessment of {IBD}: to be or not to
  be? In: Proc. IEEE EMBC. pp. 3974--3977 (2012)

\bibitem{WuCvpr20}
Wu, J., Zhang, T., Zha, Z.J., Luo, J., Zhang, Y., Wu, F.: Self-supervised
  domain-aware generative network for generalized zero-shot learning. In: The
  IEEE Conference on Computer Vision and Pattern Recognition (CVPR). pp.
  12767--12776 (June 2020)

\bibitem{Manc48}
Xian, Y., Lorenz, T., Schiele, B., Akata, Z.: Feature generating networks for
  zero-shot learning. In: In Proc. IEEE CVPR. pp. 5542--5551 (2018)

\bibitem{Xing_MICCAI19}
Xing, Y., Ge, Z., Zeng, R., Mahapatra, D., Seah, J., Law, M., Drummond, T.:
  Adversarial pulmonary pathology translation for pairwise chest x-ray data
  augmentation. In: In Proc. MICCAI. pp. 757--765 (2019)

\bibitem{CyclicGANS}
Zhu, J.Y., Park, T., Isola, P., Efros, A.A.: Unpaired image-to-image
  translation using cycle-consistent adversarial networks. In: arXiv preprint
  arXiv:1703.10593 (2017)

\bibitem{Mahapatra_MLMI15_Optic}
Zilly, J., Buhmann, J., Mahapatra, D.: Boosting convolutional filters with
  entropy sampling for optic cup and disc image segmentation from fundus
  images. In: In Proc. MLMI. pp. 136--143 (2015)

\bibitem{Zilly_CMIG_2016}
Zilly, J., Buhmann, J., Mahapatra, D.: Glaucoma detection using entropy
  sampling and ensemble learning for automatic optic cup and disc segmentation.
  In Press Computerized Medical Imaging and Graphics  \textbf{55}(1),  28--41
  (2017)

\end{thebibliography}
